\documentclass{article}

\usepackage[nonatbib,preprint]{2026template/neurips_2026}
\usepackage[numbers]{natbib}

\usepackage[export]{adjustbox}
\usepackage[ruled]{algorithm2e}
\usepackage[inline, shortlabels]{enumitem}
\usepackage[T1]{fontenc}
\usepackage{hyperref}
\usepackage{microtype}
\usepackage[framemethod=TikZ]{mdframed}
\usepackage{pifont}
\usepackage[table]{xcolor}
\usepackage{xurl}
\usepackage{placeins}
\usepackage{graphicx}
\usepackage{booktabs}
\usepackage{tabularray}
\usepackage{wrapfig}
\usepackage[most]{tcolorbox}
\usepackage{listings}
\usepackage{amsmath, amsfonts}
\usepackage{nicefrac}

\UseTblrLibrary{booktabs}

\makeatletter
\newcommand{\ssymbol}[1]{\@fnsymbol{#1}}
\newcommand{\romanNumeral}[1]{\expandafter\@slowromancap\romannumeral #1@}
\makeatother

\tcbset{
  keyfindingstyle/.style={
    enhanced,
    colback=blue!4,
    colframe=blue!50!black,
    arc=2mm,
    boxrule=0.7pt,
    left=4mm,
    right=4mm,
    top=2mm,
    bottom=2mm,
    fontupper=\small,
    breakable,
    before skip=6pt,
    after skip=6pt,
  }
}
\newtcolorbox{keyfinding}[1][]{keyfindingstyle, #1}

\usepackage{booktabs}
\usepackage{graphicx}
\usepackage{amsmath}
\usepackage{amssymb}
\usepackage{url}
\usepackage{xspace}
\usepackage{subcaption}
\usepackage{multirow}
\usepackage{makecell}
\usepackage{siunitx}
\usepackage{placeins}

\hypersetup{hidelinks}

\graphicspath{{figures/}}

\title{Beyond Encoder Accumulation: Measuring Encoder Roles in Multi-Encoder VLMs}

\author{%
  Wei Ding\thanks{Equal contribution.} \\
  Tsinghua University\\
  Beijing, China\\
  \texttt{teresading999@gmail.com} \\
  \And
  Yudong Zhang\footnotemark[1]~~\thanks{Corresponding authors.} \\
  Tsinghua University, Tencent \\
  Beijing, China \\
  \texttt{zhangyd16@mails.tsinghua.edu.cn} \\
  \AND
  Ruobing Xie \\
  Tencent \\
  Beijing, China \\
  \texttt{xrbsnowing@163.com} \\
  \And
  Xingwu Sun \\
  University of Macau \\
  Macau, China \\
  \texttt{sunxingwu01@gmail.com} \\
  \And
  Jiansheng Chen \\
  University of Science and Technology Beijing \\
  Beijing, China \\
  \texttt{jschen@ustb.edu.cn} \\
  \And
  Yu Wang\footnotemark[2] \\
  Tsinghua University \\
  Beijing, China \\
  \texttt{yu-wang@mail.tsinghua.edu.cn} \\
}

\begin{document}

\maketitle

\begin{abstract}

As foundation models scale toward fusing more heterogeneous visual streams, understanding how diverse encoders interact under joint training becomes a prerequisite for principled design. Yet large vision-language models (LVLMs) currently lack the tools to do so, and parameter-efficient encoder configurations remain hard to identify before training. To re-examine encoder roles under joint training, on the 16-benchmark Cambrian-1 suite we retrain and evaluate all 31 non-empty subsets of five common vision encoders under a unified pipeline ($\approx$20k GPU-hours total), and report three findings. First, retraining each subset from scratch reveals encoder rankings that differ from those obtained by masking encoders on a fixed checkpoint, including which encoder ranks first overall. Second, we decompose each encoder's contribution into two axes, \emph{Capacity}, the score an encoder reaches on its own, and \emph{Necessity}, the drop when it is removed from the full pool. The two axes are not interchangeable. Pairing the two highest-Capacity encoders is suboptimal, while pairing a high-Capacity anchor with an adaptive complement matches the full five-encoder model. Adding further encoders beyond this pair yields only marginal gains. Third, at fixed parameter count, per-encoder pre-projector effective rank explains the residual score variation. The strongest pairs combine an anchor whose rank survives joint training with a complement whose rank \emph{expands} under it, suggesting that higher-rank, less-collapsed projector inputs correspond to a more favorable optimization regime at the encoder–projector interface. Together, the Capacity–Necessity decomposition and the pre-projector rank analysis, along with comprehensive evaluation through retraining, expose a methodological gap in multi-encoder LVLM design, and offer concrete primitives for closing it.

\end{abstract}

\section{Introduction}
\label{sec:intro}

Large Vision-Language Models (LVLMs)~\citep{llava,zhu2023minigpt} have
made substantial progress on multimodal tasks, and combining multiple
heterogeneous vision encoders has been proposed as one route to further
gains. Eagle~\citep{shi2025eagle} is a representative example, fusing
features from ConvNeXt~\citep{liu2022convnet}, EVA-02~\citep{fang2023eva02},
CLIP~\citep{radford2021learning}, Pix2Struct~\citep{lee2022pix2struct}, and
SAM~\citep{kirillov2023segment} through channel concatenation, a recipe
also adopted by Prismatic VLMs~\citep{karamcheti2024prismaticvlms} and
subsequent channel-concat designs~\citep{jiang2024clipdinovisualencoders};
the channel-concat family is one of three fusion mechanisms surveyed
in Section~\ref{sec:related}, alongside token-sequence concatenation
and cross-attention aggregation (which Cambrian-1~\citep{tong2024cambrian1}
and BRAVE~\citep{kar2024brave} adopt). The same trend extends beyond
benchmark VLMs to vision-language-action policies~\citep{kim2024openvla},
and foundation models for world modeling and physical AI
~\citep{nvidia2025cosmos} continue to fuse heterogeneous perception
streams, making principled methodology for evaluating multi-encoder
fusion broadly relevant.

Vision encoders are much smaller than the language backbone, so adding
them is an attractive design lever if the gains are real. Some studies have attempted to conduct ablation analyses on the roles of different encoders in multi-encoder LVLMs, for example, \citet{wang2026cur} study this question by masking individual encoder
branches at inference time and measuring the resulting performance
drops, formalizing the protocol as the Conditional Utilization Rate
(CUR). Two limitations remain: this inference-time evaluation departs
from the actual training pipeline and tends to produce unreliable
conclusions, and no prior method evaluates the full set of encoder
subsets. We adopt the five-encoder family as a concrete instance to
study.
\begin{figure}[h]
  \centering
  \includegraphics[width=\linewidth]{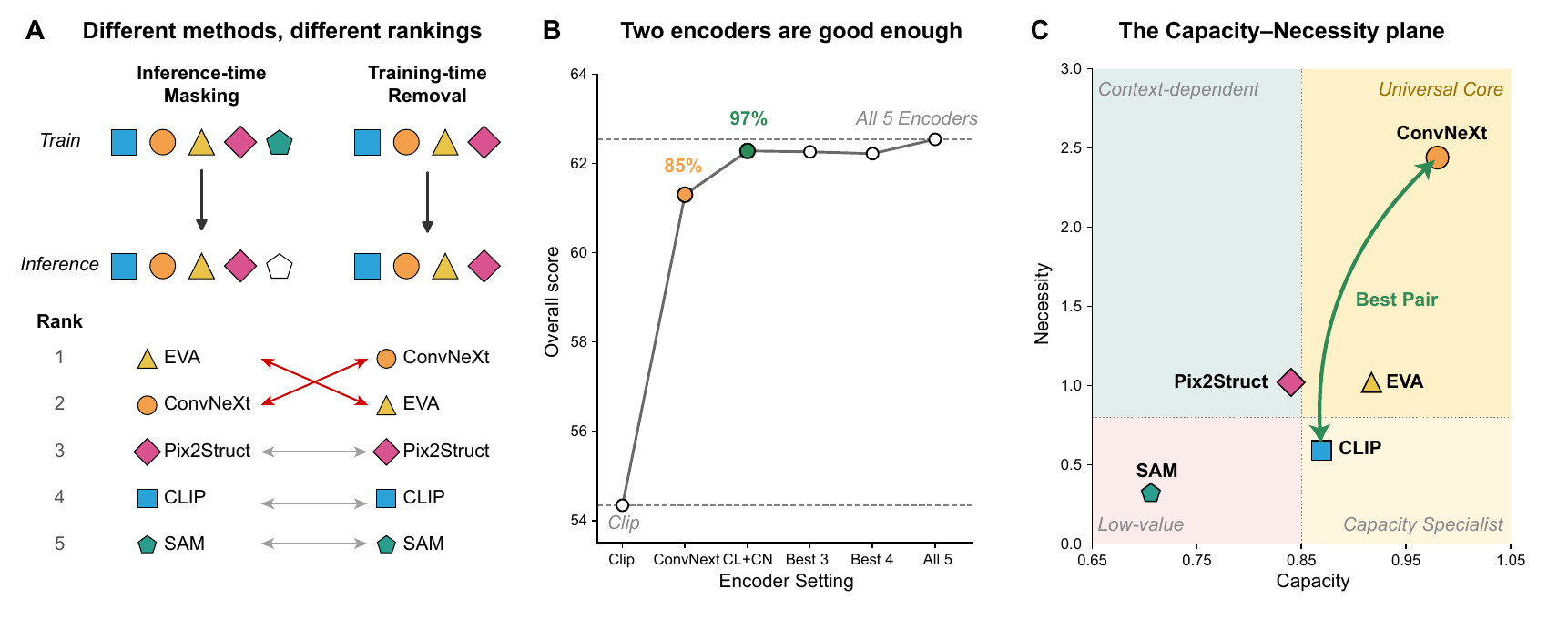}
\caption{\textbf{Paradigm preview.}
(A) IM and TR rank a different encoder first: EVA-02 under IM, ConvNeXt under TR. The two protocols also swap at rank 2, while ranks 3 to 5 agree.
(B) Best-at-$k$ overall score. With CLIP alone as the baseline and the full pool as the ceiling, ConvNeXt alone closes 85\% of the gap and CLIP+ConvNeXt closes 97\%. The third and fourth encoders add little.
(C) Capacity--Necessity plane. The five encoders fall into the four regions of the plane. The arrow points to the best two-encoder pair, ConvNeXt+CLIP, which is not the pair formed by the two highest-Capacity encoders.}
  \label{fig:main1_paradigm}
\end{figure}

\vspace{-0.5em}

We address these limitations by retraining all 31 encoder subsets
under a unified training and evaluation pipeline. Taking the
Inference-time Masking (IM) results of \citeauthor{wang2026cur} as the
reference and our Training-time Removal (TR) results as the ground
truth, we identify two systematic discrepancies. IM and TR
rank a different encoder first, and they disagree on which
two-encoder combination is best, in a way that changes the practical
recommendation. IM also shows higher cross-subset variance than TR for
most encoders, indicating that retraining each subset gives a more
stable measurement of individual encoder contributions.

Building on this protocol, three findings shape our recommendation for
compact-pool design. First, adding vision encoders yields strongly
diminishing returns: a well-chosen two-encoder model recovers nearly
all of the full-pool score, while the third and fourth encoders
contribute little, and only the fifth recovers a small Vision-Centric
margin. Second, encoder contributions decompose along two axes we
call Capacity and Necessity. A single attribution score conflates how
well an encoder performs alone with how much of its contribution is
irreplaceable by the others, and disentangling them yields a
five-class encoder taxonomy together with a non-obvious pair-selection
rule: pairing the two highest-Capacity encoders is suboptimal. Third,
per-encoder pre-projector effective rank explains residual score
variation at fixed parameter count, connecting compact-pool selection
to a representation-level signal whose change under joint training
predicts pair quality. Figure~\ref{fig:main1_paradigm} previews the
protocol contrast and the two-axis decomposition.

The contributions of this paper are summarized as follows: (1) \textbf{Attribution.} A retrain-and-remove protocol that yields
  more stable, decision-relevant rankings than inference-time masking,
  and disagrees with masking on which encoder ranks first. (2) \textbf{Decomposition.} The Capacity and Necessity axes, which
  expose substitutability that single-score attribution conflates and
  yield a Universal Core with Adaptive Complement pair-selection
  pattern that the two-highest-Capacity heuristic misses. (3) \textbf{Mechanism.} Per-encoder pre-projector effective rank as
  a representation-level selection signal independent of parameter
  count, supporting an anchor with rank-expanding complement principle
  for compact-pool design.

\section{Related Work}
\label{sec:related}

\paragraph{Multi-Encoder LVLMs.}
Multi-encoder LVLMs combine heterogeneous visual streams via three
fusion mechanisms~\citep{shi2025eagle}: token-sequence
concatenation~\citep{lin2023sphinx,fan2024mousi}, cross-attention
aggregation~\citep{li2024minigemini,tong2024cambrian1,kar2024brave,luo2024feast},
and channel-level
concatenation~\citep{karamcheti2024prismaticvlms,jiang2024clipdinovisualencoders};
\citet{tong2024cambrian1} provide the largest per-encoder benchmarking
to date. A routing line activates encoders conditionally per
input~\citep{zong2024mova,lee2024moai,zhang2025scope}, with
\citet{zhang2025scope} showing that one routed encoder can match four
fused ones. A complementary direction compresses a multi-encoder pool
into one backbone via
distillation~\citep{ranzinger2024radio,heinrich2025radiov25,cao2025movekd,wang2025hawaii},
implicitly assuming every teacher's contribution is worth preserving.
We adopt Eagle-X5~\citep{shi2025eagle} as our testbed (channel-concat
keeps token count independent of the active encoder set, making
31-subset enumeration tractable). Relative to all three lines above,
our contribution is \textbf{methodological rather than architectural}:
the 31-subset audit quantifies how much of each teacher's contribution
actually survives joint training, providing direct guidance for
distillation pipelines.

\paragraph{Encoder Attribution and Representation Analysis.}
Identifying which encoders a multi-encoder LVLM relies on after joint
training is a measurement problem. The closest precursor,
\citet{wang2026cur}, formalises this for Eagle-X5 via Conditional
Utilization Rate (CUR) and Information Gap (IG), computed by
Inference-time Masking (IM) on a fixed checkpoint; earlier inference-time
subset removal appears in BRAVE~\citep{kar2024brave}, and broader
feature-attribution
methods~\citep{heimersheim2024meanablation,wang2023ioi,sundararajan2017ig,lundberg2017shap}
share the same checkpoint-conditional counterfactual. Compact-pool
design instead requires a \emph{training-time} counterfactual: how the
model would have performed had the subset been trained from the start.
We retrain on every non-empty subset (31 models for $k{=}5$) and show
that Inference-time Masking (IM) and Training-time Removal (TR) rank a
\emph{different} encoder first; our Capacity/Necessity decomposition
then yields a five-class taxonomy and pair-selection rules the IM
protocol cannot recover. To interpret subset-level differences we
measure each encoder's pre-projector representation through its
effective rank~\citep{roy2007effectiverank}, complemented by Centred
Kernel Alignment~\citep{kornblith2019cka} as a drift-to-correctness
consistency check and a PID-based pairwise redundancy diagnostic. We
adopt a pool-conditional view of encoder roles, motivated by the
Platonic Representation Hypothesis~\citep{huh2024platonic}, which
posits that encoders may converge toward a shared statistical model
while differing in how much of it they engage.
\section{Experimental Setup}
\label{sec:setup}

\subsection{Architecture, subsets, and benchmarks}
\label{sec:setup-arch}

We build on Eagle-X5~\citep{shi2025eagle}, which pairs a Vicuna-7B
decoder~\citep{zheng2023vicuna} with five vision encoders (ConvNeXt-1024,
EVA-02-1024, CLIP-448, Pix2Struct-1024, SAM-1024). Width-aligned features
are concatenated channel-wise and mapped to the LLM input space by a
shared two-layer MLP projector; combined visual width ranges from $1024$
to $7680$ across active subsets, while projector output is fixed at
$4096$. We retrain all $2^5{-}1{=}31$ non-empty encoder subsets under a
single unified recipe, and evaluate on the 16-benchmark Cambrian-1 suite
in four families: General, Knowledge, OCR\&Chart, and Vision-Centric.
Unless stated otherwise, the overall score is the unweighted 16-task
mean and encoders are listed in Capacity-descending order
(Section~\ref{sec:capnec}). Per-encoder citations, the lattice
composition (5 singletons, 10 pairs, 10 triples, 5 quads, 1 full pool),
and the per-family benchmark lists are in
Appendix~\ref{app:setup-details}. The same 31 models support the
protocol audit (Section~\ref{sec:protocol_audit}), the Capacity and
Necessity decomposition (Section~\ref{sec:capnec}), and the per-encoder
effective rank analysis (Section~\ref{sec:mechanism}).

\subsection{Attribution protocols: IM and TR}
\label{sec:attribution-protocols}

We compare two per-encoder contribution measures. \emph{Inference-time
Masking} (IM) silences one encoder slice on a fixed full-pool
checkpoint; we use the zero-input (ZA) CUR values from
\citet{wang2026cur} as our IM reference, so any difference between IM
and TR is not confounded by reimplementation. \emph{Training-time
Removal} (TR) computes
$\mathrm{score}(F) - \mathrm{score}(F \setminus \{e\})$ from the
leave-one-out row of our retrained subsets. The two protocols answer
different counterfactuals---what the model would do if an encoder were
silenced at inference, versus what it would have learned if an encoder
had never been included---so their numerical magnitudes are not
directly commensurate; we restrict the comparison to rankings and to
the architectural choices they induce. Each subset is trained once, so
conclusions rest on ranking consistency, gap-closure patterns, and
per-family signals across the 31 models rather than pointwise
significance claims.

\begin{table}[t]
  \centering
  \caption{\textbf{Best-at-$k$ models, $k = 1, \ldots, 5$.}
  Each row reports the highest-scoring retrained subset of size $k$.}
  \label{tab:main2_bestatk}
  \scriptsize
  \resizebox{\linewidth}{!}{
\begin{tabular}{lcccccc}
\toprule
Pool (best at $k$) & $k$ & Avg & Rel. Avg & Params (M) & Throughput & Pareto \\
\midrule
ConvNeXt & 1 & 61.30 & 0.980 & 846 & 4.89 & Yes \\
CLIP + ConvNeXt & 2 & 62.28 & 0.996 & 1150 & 4.71 & Yes \\
CLIP + ConvNeXt + Pix2Struct & 3 & 62.26 & 0.996 & 1663 & 3.33 & No \\
CLIP + ConvNeXt + EVA-02 + Pix2Struct & 4 & 62.22 & 0.995 & 1966 & 2.76 & No \\
CLIP + ConvNeXt + SAM + EVA-02 + Pix2Struct & 5 & 62.54 & 1.000 & 2274 & 2.34 & Yes \\
\bottomrule
\end{tabular}
}
\end{table}

\subsection{Protocol audit: IM and TR lead to different design decisions}
\label{sec:protocol_audit}

\begin{figure}[t]
  \centering
  \includegraphics[width=0.85\linewidth]{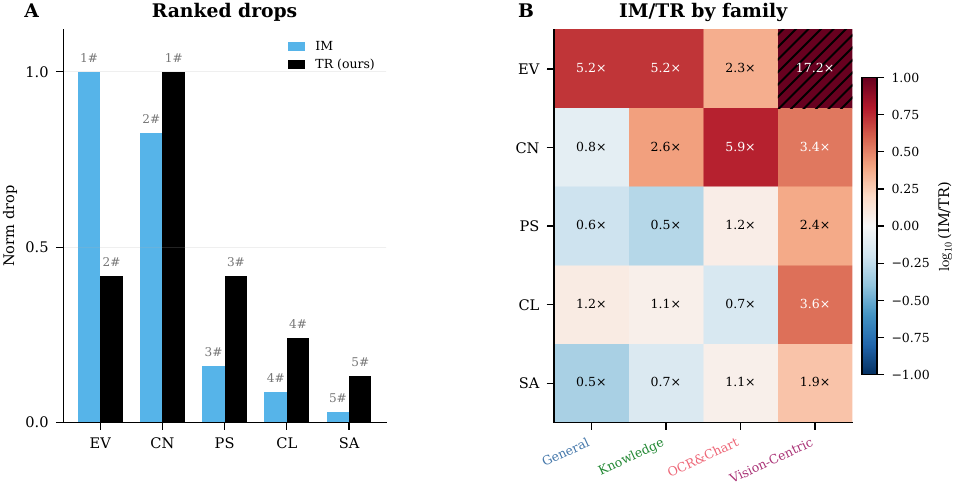}
  \caption{\textbf{Protocol audit.}
  (A) Protocol-internal normalised drops: EVA-02 is rank-1 under IM,
  ConvNeXt rank-1 under TR ($\rho{=}0.82$). EV=EVA-02, CN=ConvNeXt,
  CL=CLIP, PS=Pix2Struct, SA=SAM.
  (B) Per-encoder, per-family $\log_{10}$(IM/TR); only the largest
  outliers are annotated.}
  \label{fig:main3_audit}
\end{figure}

The two rankings broadly agree (Spearman $\rho{=}0.82$;
Figure~\ref{fig:main3_audit}A) but disagree at the most
decision-relevant position: under IM, EVA-02 ranks first and ConvNeXt
second, while under TR the order is reversed, with ConvNeXt producing
the largest leave-one-out drop. This top-anchor swap propagates into
pair selection: taking CLIP-only as a common baseline, the IM top-two
heuristic recommends ConvNeXt+EVA-02 (closing $\sim$$91\%$ of the
CLIP-to-full gap), whereas the best retrained pair is CLIP+ConvNeXt
(closing $97\%$; Table~\ref{tab:main2_bestatk}). The pair-level score
difference is small in absolute terms, but the per-family decomposition
(Figure~\ref{fig:main3_audit}B) supports treating it as a directional
design signal.

TR also yields lower across-context variance than IM for four of the
five encoders, with Pix2Struct as the only reverse case: evaluation on
a fixed checkpoint amplifies noise tied to the specific subset context,
while comparison across retrained subsets does not. IM remains useful
for quick ranking triage, but compact-pool selection requires the
training-time counterfactual that TR provides.

\section{Capacity and Necessity: Two Axes of Encoder Contribution}
\label{sec:capnec}

\subsection{Two axes from the retrained subsets}

A single attribution score conflates how well an encoder performs
alone with how much of its contribution is actually irreplaceable
under joint training. We define two axes on the 31 retrained subsets
that disentangle these modes. Capacity is the singleton score
relative to the full five-encoder model:
\[
\mathrm{Cap}(e)=\frac{\mathrm{score}(\{e\})}{\mathrm{score}(F)},
\]
where $\mathrm{score}(\cdot)$ is the 16-task average and
$F = \{$ConvNeXt, EVA-02, CLIP, Pix2Struct, SAM$\}$.

Necessity is the leave-one-out drop from the full model:
\[
\mathrm{Nec}(e)=\mathrm{score}(F)-\mathrm{score}(F\setminus\{e\}),
\]
in absolute percentage points. $\mathrm{Nec}(e)$ measures the contribution
of encoder $e$ that is not supplied by the other four when all five are
trained together.%
An encoder with high Capacity but low Necessity performs well in
isolation but is partly substitutable once the other four are present. Capacity is determined by singleton training, while Necessity depends
on the rest of the subset.

\subsection{Per-encoder values, averaged and by family}

Table~\ref{tab:taxonomy} reports per-encoder Capacity and Necessity
values; we highlight the qualitative patterns. ConvNeXt leads on both
axes; EVA-02 has high Capacity but moderate Necessity; CLIP has
moderate Capacity and the lowest Necessity among the top three.
Per-family Capacity reveals that ConvNeXt remains uniformly strong,
while CLIP, EVA-02, and Pix2Struct each show family-specific behavior
that the cross-family average hides.

The Knowledge family is a caveat: the strong LLM prior on these tasks
inflates Knowledge Capacity for every encoder, with even SAM scoring
near the full pool ceiling, suggesting visual features contribute
little on these benchmarks. We therefore exclude Knowledge from
all per-family analyses in the rest of the paper, though figures still
show it for completeness. Figure~\ref{fig:main4_capnec} summarizes the
main result; full per-encoder family profiles are in
Appendix~\ref{sec:loo-predictor-diagnostic}.

\begin{figure}[t]
  \centering
  \includegraphics[width=\linewidth]{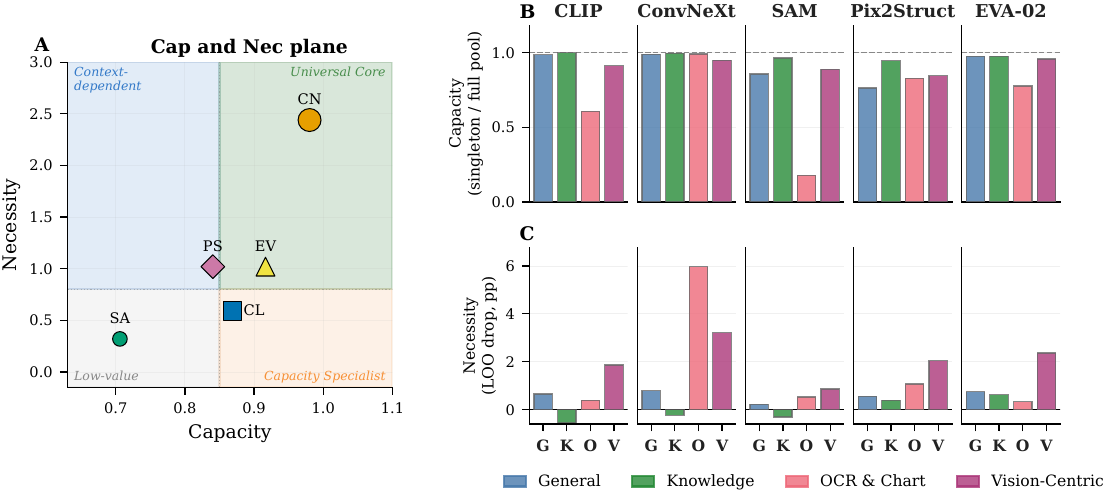}
  \caption{\textbf{Capacity and Necessity.}
  (A) The Capacity and Necessity plane. Background fills mark the four
  coarse regions (Universal Core, Context-dependent, Capacity Specialist,
  and Low-value) defined by the Cap${=}0.85$ and Nec${=}0.80$\,pp
  dotted lines; marker shape and color encode the per-encoder role
  labels in Table~\ref{tab:taxonomy}, which refine the coarse regions
  using per-family behavior and are therefore not a one-to-one map.
  (B) Per-family Capacity per encoder (singleton score normalised by
  the full pool score); the dashed line marks the full-model ceiling.
  (C) Per-family Necessity per encoder (leave-one-out drop in pp), with
  the full $y$-range shown. Knowledge bars (green, K) reflect the LLM
  prior rather than visual attribution; the interpretive focus is on
  the other three families.}
  \label{fig:main4_capnec}
\end{figure}

\subsection{Per-encoder roles and pair selection}

Each encoder receives a role label combining its Capacity and Necessity
values with per-family behavior (Table~\ref{tab:taxonomy}). ConvNeXt
is the \emph{Universal Core}, uniformly strong across families. CLIP
is an \emph{Adaptive Complement}: moderate Capacity, but the only
encoder whose effective rank grows under joint training
(Section~\ref{sec:mechanism}). EVA-02 is a \emph{Vision-Centric Specialist}
whose Necessity depends strongly on the rest of the subset, and the
best alternative primary for Vision-Centric tasks. Pix2Struct is a
\emph{Niche Specialist} concentrated on OCR\&Chart, and SAM is
\emph{Replaceable}.

The best two-encoder model combines CLIP with ConvNeXt rather than
ConvNeXt with EVA-02, even though EVA-02 has higher singleton Capacity
than CLIP; per-pair overall and per-family scores for all 10
two-encoder subsets are listed in Appendix Table~\ref{tab:allpair}.
Selecting the two highest-Capacity encoders is therefore not optimal,
even within this five-encoder set. The rule consistent with our data
is instead to combine a high-Capacity primary encoder with a second
encoder that is minimally redundant with the first under joint
training. Section~\ref{sec:mechanism} measures this non-redundancy via
the second encoder's effective rank growth in the trained pair.

\begin{table}[t]
  \centering
  \caption{\textbf{Per-encoder role labels under the Capacity and
  Necessity decomposition.}}
  \label{tab:taxonomy}
  \scriptsize
  \resizebox{\linewidth}{!}{\begin{tabular}{lcccc}
\toprule
Encoder & Cap & Nec & Role label & Deployment Prescription \\
\midrule
ConvNeXt & 0.980 & 2.44 & \textbf{Universal Core} & Default anchor; positive presence effect across families. \\
CLIP & 0.869 & 0.59 & \textbf{Adaptive Complement} & Add to anchor; joint training expands rank. \\
EVA-02 & 0.917 & 1.02 & \textbf{Vision-Centric Specialist} & Alternative anchor for vision-centric deployment. \\
Pix2Struct & 0.840 & 1.02 & \textbf{Niche Specialist} & Add for OCR-heavy deployment. \\
SAM & 0.706 & 0.32 & \textbf{Replaceable} & First removal candidate in broad pool. \\
\bottomrule
\end{tabular}
}
\end{table}

\subsection{Marginal contribution across subset sizes}
\label{sec:marginal}

The marginal contribution at size $k$ is
$\Delta_k = \mathrm{best}(k) - \mathrm{best}(k-1)$, where
$\mathrm{best}(k)$ is the highest-scoring subset of $k$ encoders
(the optimum at each $k$, not a greedy extension of
$\mathrm{best}(k-1)$).

\begin{figure}[t]
  \centering
  \includegraphics[width=\linewidth]{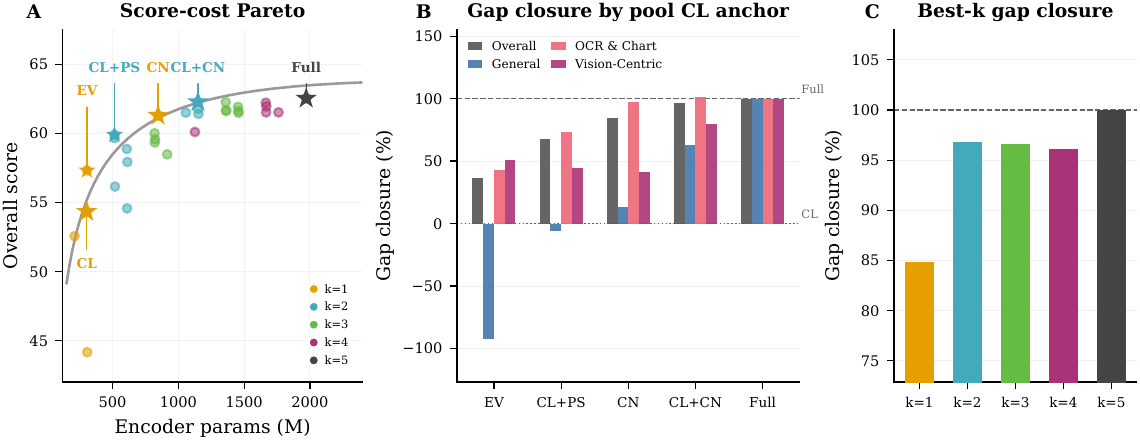}
\caption{\textbf{Two encoders are sufficient for most of the full-pool
score.}
\textbf{(A) Pareto frontier.} Overall score against encoder parameter
count for all 31 subsets, colored by size $k$. ConvNeXt alone, the
pair of CLIP and ConvNeXt, and the full pool sit on the Pareto
frontier; the three-encoder and four-encoder pools below them are
dominated. Per-subset parameter counts and throughput are listed in
Appendix Table~\ref{tab:compute_pareto}.
\textbf{(B) Per-family gap closure} for representative pools (CLIP as
common baseline). The pair of CLIP and ConvNeXt closes at least
$95\%$ General and OCR\,\&\,Chart, but Vision-Centric
keeps benefiting up to the full pool. Knowledge omitted (denominator
near zero).
\textbf{(C) Best-at-$k$ gap closure.} Most of the gap is closed at
$k{=}2$; $k{=}3$ and $k{=}4$ plateau; $k{=}5$ recovers a small
additional margin.}
  \label{fig:main5_pairs}
\end{figure}

Adding CLIP to ConvNeXt yields the largest marginal gain. The third
and fourth encoders contribute essentially zero
(Figure~\ref{fig:main5_pairs}C), indicating saturation rather than
monotonic gain. The full five-encoder model recovers a small
Vision-Centric margin over the best four-encoder subset, which we
report as a family-specific direction rather than a significance
claim.

CLIP combined with ConvNeXt closes $97\%$ of the CLIP-to-full gap on
the overall score; on OCR\&Chart the pair actually matches or slightly
exceeds the full model, indicating early saturation on this family.
Vision-Centric does not saturate until $k{=}5$. (Knowledge is omitted
from Figure~\ref{fig:main5_pairs}B because its CLIP-to-full denominator
is near zero.) The number of encoders to include is therefore set by
the most demanding task family in the target workload, and
Section~\ref{sec:mechanism} asks what these saturating subsets share
at the representation level.

The right choice of encoders therefore depends jointly on the
available pool and the target task family: ConvNeXt alone is the
strongest singleton, CLIP combined with ConvNeXt is the best pair,
and additional encoders are justified mainly by Vision-Centric
requirements. Encoder parameter count remains a useful but coarse
predictor; Section~\ref{sec:mechanism} attributes the residual
variation at fixed parameter count to per-encoder pre-projector
effective rank.

\section{A Rank-Based View of Encoder Composition}
\label{sec:mechanism}

Sections~\ref{sec:protocol_audit} and~\ref{sec:capnec} established
\emph{which} encoder pools win and what their roles are; we now ask
\emph{what} they share at the representation level. For each retrained
checkpoint we hook the shared projector and, for every active encoder,
compute the \emph{effective rank} of its own channel slice of the
projector input
($r_{\mathrm{eff}}(X) = \exp\!\left(-\sum_i p_i \log p_i\right)$ with
$p_i = s_i^2 / \sum_j s_j^2$~\citep{roy2007effectiverank}). We measure
on the projector input rather than its output because the
post-projector $4096$-dim space is shared and cannot be cleanly
decomposed back to individual encoders. Three regimes emerge as pool
size grows, each with a different operative rank quantity
(Section~\ref{sec:rank_three_regimes}); rank does not displace
parameter count as a global predictor of score, but it captures
representational structure that parameter count alone misses
(Section~\ref{sec:rank_vs_params}).

\subsection{Three regimes}
\label{sec:rank_three_regimes}

\begin{figure}[t]
  \centering
  \includegraphics[width=\linewidth]{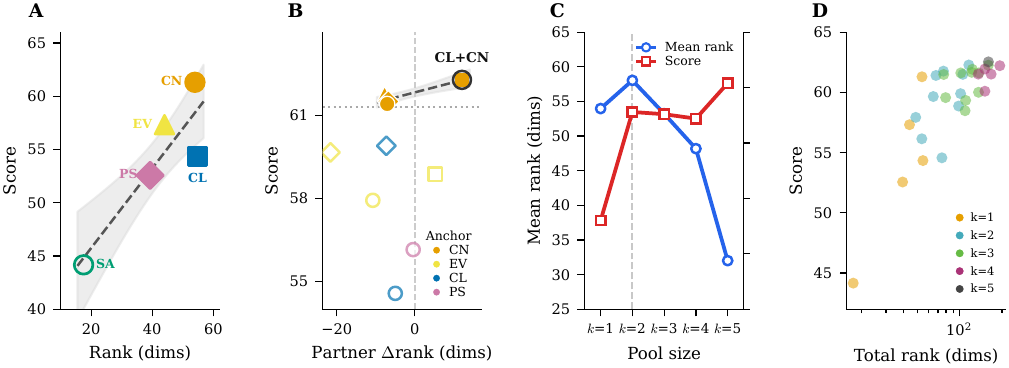}
  \caption{\textbf{Per-encoder pre-projector effective rank tracks
    score across three regimes.}
    \textbf{(A)} Singleton rank predicts singleton score
    (Pearson $r{=}0.89$).
    \textbf{(B)} Within ConvNeXt-anchored pairs, partner
    $\Delta$rank (the complement's rank in the pair minus its
    singleton baseline) tracks pair score. CLIP shows the only
    substantial rank expansion under joint training, and
    CLIP+ConvNeXt tops the pair tier.
    \textbf{(C)} Best-at-$k$ trajectory: overall score plateaus at the
    cardinality where mean rank stops growing.
    \textbf{(D)} Pool-level: score against $\sum_e r_e$ (Pearson
    $r{=}0.67$). Pools at similar aggregate rank diverge by whether
    the budget is text-aligned or fragmented across mismatched
    objectives.}
  \label{fig:rank_scaling}
\end{figure}

The headline finding is that the predictor of score \emph{shifts} as
the pool grows: from a singleton's own rank, to the complement's rank
\emph{change} under joint training, to a saturating budget shared
across encoders (Figure~\ref{fig:rank_scaling}).

\textbf{Single encoder: rank predicts score.} The projector input is
one encoder's slice, so per-encoder rank reads out representational
diversity in that encoder's native subspace; the correspondence to
singleton score is tight (Pearson $r{=}0.89$, Panel A). Capacity
(Section~\ref{sec:capnec}) is essentially the score axis of this
scatter normalised by the full-pool ceiling.

\textbf{Two encoders: the operative quantity is $\Delta$rank, not
the singleton value.} Joint training begins to reshape
representations, and the predictor switches from the singleton level
to its change. Within ConvNeXt-anchored pairs, partner $\Delta$rank
tracks pair score, and the complement whose rank \emph{expands} under
joint training (CLIP) wins---even though the alternative complement
(EVA-02) has higher singleton Capacity. What matters at the pair
level is not the level of rank an encoder brings, but whether that
rank survives or grows once it shares the projector with another
encoder.

\textbf{Multiple encoders: the projector budget saturates.} The
shared projector compresses a varying input width into a fixed
$4096$-dim output, so as more encoders share that budget, joint
training is forced to push most encoders' slices toward lower-rank
representations: the mean per-encoder rank drops by several
dimensions per addition
(Appendix~Table~\ref{tab:rank-vs-pool}). This supplies the
representation-level mechanism behind the saturation in
Section~\ref{sec:capnec}: best-at-$k$ score plateaus at the same
cardinality where mean rank stops growing.

At the pool level (Panel~D), score correlates with total per-encoder
effective rank at $r{=}0.67$, with a ceiling near the two-encoder
saturation point. Pools at similar aggregate rank can sit near the
frontier or well below it depending on whether the rank budget is
text-aligned or fragmented across mismatched objectives. Rank
quantity is therefore necessary but not sufficient: the budget must
also be allocated across encoders whose representations remain
compatible under joint training.

\subsection{Rank versus parameter count}
\label{sec:rank_vs_params}

These rank patterns are not simply a stand-in for parameter count,
though the two carry partly overlapping information. The cleanest
isolation is the matched-parameter singleton control: CLIP, EVA-02,
and SAM all have roughly $305$\,M parameters but span a wide range of
effective rank, and their singleton scores differ by more than ten
points---a spread tracked by rank but not by parameter count. Within
a fixed cardinality, parameters remain the stronger residual
predictor. We therefore use rank as a complementary descriptive lens
grounded in representation geometry, not as an independent scaling
dimension.

\subsection{Task families differ in rank demand}
\label{sec:rank_demand}

\begin{figure}[t]
  \centering
  \includegraphics[width=0.85\linewidth]{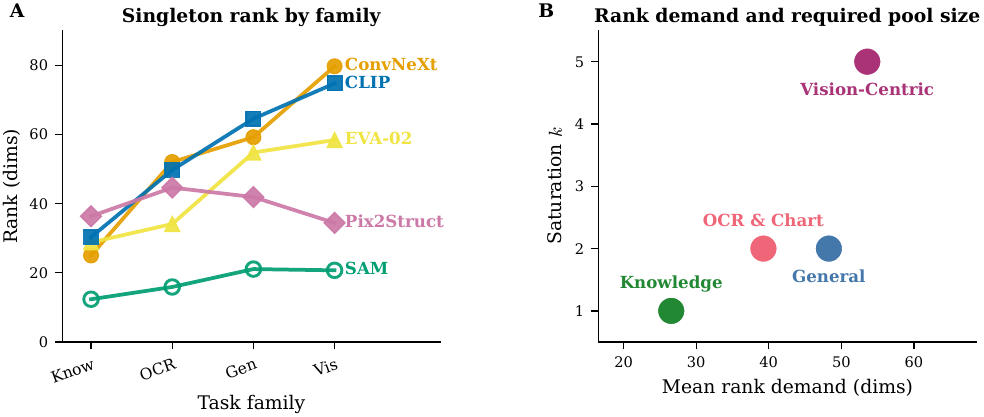}
  \caption{\textbf{Task families differ in rank demand.}
    \textbf{(A)} Singleton effective rank by family. Knowledge engages
    little rank across all encoders (LLM-prior dominated);
    Vision-Centric demands the most.
    \textbf{(B)} Family rank demand against saturation $k$ (smallest
    pool whose best-at-$k$ family score reaches $99\%$ of the
    full-pool family score). Vision-Centric is the only family whose
    demand exceeds every singleton's budget.}
  \label{fig:rank_demand}
\end{figure}
The family-asymmetric saturation observed in
Section~\ref{sec:capnec}---OCR\&Chart saturating with two encoders,
Vision-Centric continuing through five---has a representation-level
counterpart in rank demand. Singleton pre-projector rank is low on
\textsc{knowledge} (LLM-prior dominated) and highest on
\textsc{vision-centric}; only \textsc{convnext} and \textsc{clip}
have the headroom to reach high rank there, while \textsc{eva-02},
\textsc{pix2struct}, and \textsc{sam} plateau lower. Rank demand and
family saturation therefore co-vary
(Figure~\ref{fig:rank_demand}\,B): families met by a single
high-rank encoder (\textsc{ocr\&chart}, \textsc{general}) saturate
with two encoders, while \textsc{vision-centric}---the only family
whose demand exceeds any singleton's budget---keeps benefiting from
larger pools. The cardinality decision in
Section~\ref{sec:capnec} thus has a clean representation-level
reading: aggregate enough rank to clear the highest family-level
demand, and no more.

\subsection{Anchor with rank-expanding complement}
\label{sec:rank_synthesis}

Putting the three regimes together: strong compact pools combine
(i) a high-Capacity anchor whose own rank survives joint training
intact, and (ii) a complement whose rank \emph{grows} under joint
training, expanding its projector-input subspace rather than
compressing it. In our five-encoder pool, \textsc{convnext} is the
only encoder satisfying (i)---its pre-projector rank shifts only
marginally across its four pairings---and \textsc{clip} is the only
one satisfying (ii): partner $\Delta$rank when CLIP serves as the
complement is positive and substantially larger than for any other
encoder (Appendix~Table~\ref{tab:rank-vs-pool}). No other encoder
satisfies both, which is why CLIP+ConvNeXt sits alone at the top of
the pair tier.

This is a descriptive heuristic for the Eagle-X5 pool, not a derived
selection rule for unseen pools; whether rank expansion is a generic
property of contrastively pretrained, text-aligned complements is
left for future work. Sample-level drift-to-correctness and IM-flip
negative-control checks
(Appendix~\ref{sec:appendix-extra-diagnostics}) are directionally
consistent with the synthesis.
\section{Discussion}
\label{sec:discussion}

\paragraph{Encoder roles are properties of joint training.}
In a multi-encoder LVLM, an encoder's contribution is a property of
joint training, not of the encoder in isolation. IM and TR rank a
different encoder first (Section~\ref{sec:protocol_audit}); the best
two-encoder pool combines an anchor with an adaptive complement
rather than the two highest-Capacity encoders
(Section~\ref{sec:capnec}); and pair quality is predicted by the
complement's rank \emph{change} under joint training, not by its
singleton value (Section~\ref{sec:mechanism}).

\paragraph{Implications for multi-encoder distillation.}
The Capacity and Necessity decomposition together with the
rank-budget view predicts when distillation
methods~\citep{ranzinger2024radio,cao2025movekd,wang2025hawaii}
should help. Compressing a Replaceable encoder should preserve
performance; a Niche Specialist preserves its family only if the
student carries that capability. The Adaptive Complement is the hard
case: its contribution comes from rank that \emph{grows} under joint
training, an encoder--projector interaction a frozen distilled
student cannot reproduce.

\paragraph{CKA-based representational diagnostics.}
Three supplementary CKA analyses reinforce the main findings:
pretrain-to-finetune drift on correctly answered queries is lower
for CLIP and EVA-02 than for Pix2Struct and SAM
(Appendix~\ref{sec:appendix-extra-diagnostics}), matching their
contrasting roles; projector CKA between retrained models grows
with the number of shared encoders
(Appendix~\ref{sec:appendix-cka-convergence}), confirming that pool
composition rather than random initialisation drives the projector's
fusion geometry; and a proxy PID decomposition shows pairwise
complementarity is largely additive
(Appendix~\ref{sec:appendix-pid}), consistent with encoders
contributing approximately independent subspace slices.

\paragraph{Rank and optimisation geometry.}
The pattern of best pairs combining an anchor with a rank-expanding
complement (Section~\ref{sec:rank_synthesis}) admits a
representation-level reading: a higher-rank projector input may
correspond to a more well-conditioned forward signal, in line with
prior work connecting higher effective rank of intermediate
representations to more favourable optimisation
dynamics~\citep{daneshmand2020bnrank,feng2022rankdiminishing}; the
same geometric intuition appears on the optimiser side in recent
work on orthogonalised gradient
updates~\citep{jordan2024muon,liu2025muon}. We do not directly
measure optimiser-level quantities, so this remains a
representation-level correlate left for future investigation.

\paragraph{A methodological view, and what future fusion architectures will need.}
The findings above expose a methodological gap in multi-encoder
LVLM design: how to attribute, decompose, and ablate encoder
contributions in ways that reflect joint training rather than
fixed-checkpoint approximations. As foundation models for world
modeling~\citep{nvidia2025cosmos} and self-supervised video world
models~\citep{assran2025vjepa2} fuse more heterogeneous perception
streams, full enumeration retraining becomes infeasible, and the
open question becomes how to estimate Capacity, Necessity, and
projector-input rank dynamics with minimal retraining---for
instance via lightweight surrogates from pretrained encoders'
pre-projector geometry (mutual rank, CKA, PID), calibrated against
a reference like ours. Scope and limitations of the present study
are discussed in
Appendix~\ref{sec:appendix-scope-limitations}.
\section{Conclusion}
\label{sec:conclusion}

This paper conducts retraining-based ablation experiments on encoder combinations in multi-encoder LVLMs to analyze the functional roles of individual encoders. Meanwhile, from the perspectives of capacity, necessity, and effective rank, it further proposes practical strategies for encoder selection.
Future foundation models for world modeling and embodied perception will inevitably fuse heterogeneous perceptual information flows. Our method is not only applicable to multi-encoder LVLMs, but also contributes to the development and optimization of future world modeling models.

\clearpage
\FloatBarrier

\bibliographystyle{plainnat}
\bibliography{references}

@inproceedings{llava,
  author    = {Liu, Haotian and Li, Chunyuan and Wu, Qingyang and Lee, Yong Jae},
  title     = {Visual Instruction Tuning},
  booktitle = {Advances in Neural Information Processing Systems (NeurIPS)},
  year      = {2023},
}

@article{zhu2023minigpt,
  title   = {{MiniGPT-4}: Enhancing Vision-Language Understanding with Advanced Large Language Models},
  author  = {Zhu, Deyao and Chen, Jun and Shen, Xiaoqian and Li, Xiang and Elhoseiny, Mohamed},
  journal = {arXiv preprint arXiv:2304.10592},
  year    = {2023},
}

@inproceedings{zheng2023vicuna,
  title     = {Judging {LLM}-as-a-Judge with {MT-Bench} and Chatbot Arena},
  author    = {Zheng, Lianmin and Chiang, Wei-Lin and Sheng, Ying and Zhuang, Siyuan and Wu, Zhanghao and Zhuang, Yonghao and Lin, Zi and Li, Zhuohan and Li, Dacheng and Xing, Eric P. and Zhang, Hao and Gonzalez, Joseph E. and Stoica, Ion},
  booktitle = {Advances in Neural Information Processing Systems (NeurIPS), Datasets and Benchmarks Track},
  year      = {2023},
}

@inproceedings{liu2022convnet,
  title     = {A {ConvNet} for the 2020s},
  author    = {Liu, Zhuang and Mao, Hanzi and Wu, Chao-Yuan and Feichtenhofer, Christoph and Darrell, Trevor and Xie, Saining},
  booktitle = {IEEE/CVF Conference on Computer Vision and Pattern Recognition (CVPR)},
  pages     = {11976--11986},
  year      = {2022},
}

@inproceedings{radford2021learning,
  title     = {Learning Transferable Visual Models From Natural Language Supervision},
  author    = {Radford, Alec and Kim, Jong Wook and Hallacy, Chris and Ramesh, Aditya and Goh, Gabriel and Agarwal, Sandhini and Sastry, Girish and Askell, Amanda and Mishkin, Pamela and Clark, Jack and Krueger, Gretchen and Sutskever, Ilya},
  booktitle = {International Conference on Machine Learning (ICML)},
  pages     = {8748--8763},
  year      = {2021},
}

@article{assran2025vjepa2,
  title   = {{V-JEPA 2}: Self-Supervised Video Models Enable Understanding, Prediction and Planning},
  author  = {Assran, Mahmoud and Bardes, Adrien and Fan, David and Garrido, Quentin and Howes, Russell and Komeili, Mojtaba and Muckley, Matthew and Rizvi, Ammar and Roberts, Claire and Sinha, Koustuv and others},
  journal = {arXiv preprint arXiv:2506.09985},
  year    = {2025},
}

@inproceedings{kirillov2023segment,
  title     = {Segment Anything},
  author    = {Kirillov, Alexander and Mintun, Eric and Ravi, Nikhila and Mao, Hanzi and Rolland, Chloe and Gustafson, Laura and Xiao, Tete and Whitehead, Spencer and Berg, Alexander C. and Lo, Wan-Yen and Doll{\'a}r, Piotr and Girshick, Ross},
  booktitle = {IEEE/CVF International Conference on Computer Vision (ICCV)},
  pages     = {4015--4026},
  year      = {2023},
}

@inproceedings{lee2022pix2struct,
  title     = {{Pix2Struct}: Screenshot Parsing as Pretraining for Visual Language Understanding},
  author    = {Lee, Kenton and Joshi, Mandar and Turc, Iulia and Hu, Hexiang and Liu, Fangyu and Eisenschlos, Julian and Khandelwal, Urvashi and Shaw, Peter and Chang, Ming-Wei and Toutanova, Kristina},
  booktitle = {International Conference on Machine Learning (ICML)},
  year      = {2023},
  note      = {arXiv:2210.03347},
}

@article{fang2023eva02,
  title   = {{EVA-02}: A Visual Representation for Neon Genesis},
  author  = {Fang, Yuxin and Sun, Quan and Wang, Xinggang and Huang, Tiejun and Wang, Xinlong and Cao, Yue},
  journal = {Image and Vision Computing},
  volume  = {149},
  pages   = {105171},
  year    = {2024},
  note    = {arXiv:2303.11331},
}

@misc{fan2024mousi,
  title         = {{MouSi}: Poly-Visual-Expert Vision-Language Models},
  author        = {Fan, Xiaoran and Ji, Tao and Jiang, Changhao and Li, Shuo and Jin, Senjie and Song, Sirui and Wang, Junke and Hong, Boyang and Chen, Lu and Zheng, Guodong and others},
  year          = {2024},
  eprint        = {2401.17221},
  archivePrefix = {arXiv},
  primaryClass  = {cs.CV},
}

@inproceedings{kar2024brave,
  title     = {{BRAVE}: Broadening the Visual Encoding of Vision-Language Models},
  author    = {Kar, O{\u g}uzhan Fatih and Tonioni, Alessio and Poklukar, Petra and Kulshrestha, Achin and Zamir, Amir and Tombari, Federico},
  booktitle = {European Conference on Computer Vision (ECCV)},
  year      = {2024},
  note      = {Oral; arXiv:2404.07204},
}

@InProceedings{lin2023sphinx,
author="Lin, Ziyi
and Liu, Dongyang
and Zhang, Renrui
and Gao, Peng
and Qiu, Longtian
and Xiao, Han
and Qiu, Han
and Shao, Wenqi
and Chen, Keqin
and Han, Jiaming
and Huang, Siyuan
and Zhang, Yichi
and He, Xuming
and Qiao, Yu
and Li, Hongsheng",
editor="Leonardis, Ale{\v{s}}
and Ricci, Elisa
and Roth, Stefan
and Russakovsky, Olga
and Sattler, Torsten
and Varol, G{\"u}l",
title="SPHINX: A Mixer of Weights, Visual Embeddings and Image Scales for Multi-modal Large Language Models",
booktitle="Computer Vision -- ECCV 2024",
year="2025",
publisher="Springer Nature Switzerland",
address="Cham",
pages="36--55",
isbn="978-3-031-73033-7"
}

@inproceedings{karamcheti2024prismaticvlms,
  title     = {Prismatic {VLMs}: Investigating the Design Space of Visually-Conditioned Language Models},
  author    = {Karamcheti, Siddharth and Nair, Suraj and Balakrishna, Ashwin and Liang, Percy and Kollar, Thomas and Sadigh, Dorsa},
  booktitle = {International Conference on Machine Learning (ICML)},
  year      = {2024},
  note      = {arXiv:2402.07865},
}

@misc{jiang2024clipdinovisualencoders,
  title         = {From {CLIP} to {DINO}: Visual Encoders Shout in Multi-modal Large Language Models},
  author        = {Jiang, Dongsheng and Liu, Yuchen and Liu, Songlin and Zhao, Jin'e and Zhang, Hao and Gao, Zhen and Zhang, Xiaopeng and Li, Jin and Xiong, Hongkai},
  year          = {2024},
  eprint        = {2310.08825},
  archivePrefix = {arXiv},
  primaryClass  = {cs.CV},
}

@inproceedings{shi2025eagle,
  title     = {Eagle: Exploring The Design Space for Multimodal {LLMs} with Mixture of Encoders},
  author    = {Shi, Min and Liu, Fuxiao and Wang, Shihao and Liao, Shijia and Radhakrishnan, Subhashree and Zhao, Yilin and Huang, De-An and Yin, Hongxu and Sapra, Karan and Yacoob, Yaser and Shi, Humphrey and Catanzaro, Bryan and Tao, Andrew and Kautz, Jan and Yu, Zhiding and Liu, Guilin},
  booktitle = {International Conference on Learning Representations (ICLR)},
  year      = {2025},
  note      = {Spotlight; arXiv:2408.15998},
}

@article{tong2024cambrian1,
  title={Cambrian-1: A fully open, vision-centric exploration of multimodal llms},
  author={Tong, Shengbang and Brown, Ellis and Wu, Penghao and Woo, Sanghyun and Middepogu, Manoj and Akula, Sai C and Yang, Jihan and Yang, Shusheng and Iyer, Adithya and Pan, Xichen and others},
  journal={Advances in Neural Information Processing Systems},
  volume={37},
  pages={87310--87356},
  year={2024}
}

@misc{li2024minigemini,
  title         = {{Mini-Gemini}: Mining the Potential of Multi-modality Vision Language Models},
  author        = {Li, Yanwei and Zhang, Yuechen and Wang, Chengyao and Zhong, Zhisheng and Chen, Yixin and Chu, Ruihang and Liu, Shaoteng and Jia, Jiaya},
  year          = {2024},
  eprint        = {2403.18814},
  archivePrefix = {arXiv},
  primaryClass  = {cs.CV},
}

@inproceedings{luo2024feast,
  title     = {Feast Your Eyes: Mixture-of-Resolution Adaptation for Multimodal Large Language Models},
  author    = {Luo, Gen and Zhou, Yiyi and Zhang, Yuxin and Zheng, Xiawu and Sun, Xiaoshuai and Ji, Rongrong},
  booktitle = {International Conference on Learning Representations (ICLR)},
  year      = {2025},
  note      = {arXiv:2403.03003},
}

@inproceedings{zong2024mova,
  title     = {{MoVA}: Adapting Mixture of Vision Experts to Multimodal Context},
  author    = {Zong, Zhuofan and Ma, Bingqi and Shen, Dazhong and Song, Guanglu and Shao, Hao and Jiang, Dongzhi and Li, Hongsheng and Liu, Yu},
  booktitle = {Advances in Neural Information Processing Systems (NeurIPS)},
  year      = {2024},
  note      = {arXiv:2404.13046},
}

@inproceedings{lee2024moai,
  title     = {{MoAI}: Mixture of All Intelligence for Large Language and Vision Models},
  author    = {Lee, Byung-Kwan and Park, Beomchan and Kim, Chae Won and Ro, Yong Man},
  booktitle = {European Conference on Computer Vision (ECCV)},
  year      = {2024},
  note      = {arXiv:2403.07508},
}

@misc{zhang2025scope,
  title         = {{SCOPE}: Selective Cross-modal Orchestration of Visual Perception Experts},
  author        = {Zhang, Tianyu and Wang, Suyuchen and Wang, Chao and Rodriguez, Juan and Masry, Ahmed and Jian, Xiangru and Bengio, Yoshua and Taslakian, Perouz},
  year          = {2025},
  eprint        = {2510.12974},
  archivePrefix = {arXiv},
  primaryClass  = {cs.CV},
}

@inproceedings{ranzinger2024radio,
  title     = {{AM-RADIO}: Agglomerative Vision Foundation Model -- Reduce All Domains Into One},
  author    = {Ranzinger, Mike and Heinrich, Greg and Kautz, Jan and Molchanov, Pavlo},
  booktitle = {IEEE/CVF Conference on Computer Vision and Pattern Recognition (CVPR)},
  pages     = {12490--12500},
  year      = {2024},
  note      = {arXiv:2312.06709},
}

@misc{heinrich2025radiov25,
      title={RADIOv2.5: Improved Baselines for Agglomerative Vision Foundation Models}, 
      author={Greg Heinrich and Mike Ranzinger and Hongxu and Yin and Yao Lu and Jan Kautz and Andrew Tao and Bryan Catanzaro and Pavlo Molchanov},
      year={2025},
      eprint={2412.07679},
      archivePrefix={arXiv},
      primaryClass={cs.CV},
      url={https://arxiv.org/abs/2412.07679}, 
}

@inproceedings{cao2025movekd,
  title     = {{MoVE-KD}: Knowledge Distillation for {VLMs} with Mixture of Visual Encoders},
  author    = {Cao, Jiajun and Zhang, Yuan and Huang, Tao and Lu, Ming and Zhang, Qizhe and An, Ruichuan and Ma, Ningning and Zhang, Shanghang},
  booktitle = {IEEE/CVF Conference on Computer Vision and Pattern Recognition (CVPR)},
  year      = {2025},
  note      = {arXiv:2501.01709},
}

@inproceedings{wang2025hawaii,
  title     = {{HAWAII}: Hierarchical Visual Knowledge Transfer for Efficient Vision-Language Models},
  author    = {Wang, Yimu and Azadani, Mozhgan Nasr and Sedwards, Sean and Czarnecki, Krzysztof},
  booktitle = {Advances in Neural Information Processing Systems (NeurIPS)},
  year      = {2025},
  note      = {arXiv:2506.19072},
}

@inproceedings{wang2026cur,
  title     = {Investigating Redundancy in Multimodal Large Language Models with Multiple Vision Encoders},
  author    = {Wang, Yizhou and Mao, Song and Chen, Yang and Shen, Yufan and Cai, Pinlong and Wang, Ding and Yan, Guohang and Yu, Zhi and Yan, Yinqiao and Hu, Xuming and Shi, Botian},
  booktitle = {International Conference on Learning Representations (ICLR)},
  year      = {2026},
  note      = {arXiv:2507.03262},
}

@misc{heimersheim2024meanablation,
  title         = {How to use and interpret activation patching},
  author        = {Heimersheim, Stefan and Nanda, Neel},
  year          = {2024},
  eprint        = {2404.15255},
  archivePrefix = {arXiv},
  primaryClass  = {cs.LG},
}

@inproceedings{wang2023ioi,
  title     = {Interpretability in the Wild: A Circuit for Indirect Object Identification in {GPT-2} Small},
  author    = {Wang, Kevin and Variengien, Alexandre and Conmy, Arthur and Shlegeris, Buck and Steinhardt, Jacob},
  booktitle = {International Conference on Learning Representations (ICLR)},
  year      = {2023},
  note      = {arXiv:2211.00593},
}

@inproceedings{sundararajan2017ig,
  title     = {Axiomatic Attribution for Deep Networks},
  author    = {Sundararajan, Mukund and Taly, Ankur and Yan, Qiqi},
  booktitle = {International Conference on Machine Learning (ICML)},
  year      = {2017},
  note      = {arXiv:1703.01365},
}

@inproceedings{lundberg2017shap,
  title     = {A Unified Approach to Interpreting Model Predictions},
  author    = {Lundberg, Scott M. and Lee, Su-In},
  booktitle = {Advances in Neural Information Processing Systems (NeurIPS)},
  year      = {2017},
  note      = {arXiv:1705.07874},
}

@inproceedings{kornblith2019cka,
  title     = {Similarity of Neural Network Representations Revisited},
  author    = {Kornblith, Simon and Norouzi, Mohammad and Lee, Honglak and Hinton, Geoffrey},
  booktitle = {International Conference on Machine Learning (ICML)},
  year      = {2019},
  note      = {arXiv:1905.00414},
}

@inproceedings{roy2007effectiverank,
  title     = {The Effective Rank: A Measure of Effective Dimensionality},
  author    = {Roy, Olivier and Vetterli, Martin},
  booktitle = {European Signal Processing Conference (EUSIPCO)},
  pages     = {606--610},
  year      = {2007},
}

@inproceedings{huh2024platonic,
  title     = {The Platonic Representation Hypothesis},
  author    = {Huh, Minyoung and Cheung, Brian and Wang, Tongzhou and Isola, Phillip},
  booktitle = {International Conference on Machine Learning (ICML)},
  year      = {2024},
  note      = {arXiv:2405.07987},
}

@article{kim2024openvla,
  title   = {{OpenVLA}: An Open-Source Vision-Language-Action Model},
  author  = {Kim, Moo Jin and Pertsch, Karl and Karamcheti, Siddharth and Xiao, Ted and Balakrishna, Ashwin and Nair, Suraj and Rafailov, Rafael and Foster, Ethan and Lam, Grace and Sanketi, Pannag and others},
  journal = {arXiv preprint arXiv:2406.09246},
  year    = {2024},
}

@article{nvidia2025cosmos,
  title   = {Cosmos World Foundation Model Platform for Physical {AI}},
  author  = {{NVIDIA}},
  journal = {arXiv preprint arXiv:2501.03575},
  year    = {2025},
}

@inproceedings{daneshmand2020bnrank,
  title     = {Batch Normalization Provably Avoids Rank Collapse for Randomly Initialised Deep Networks},
  author    = {Daneshmand, Hadi and Kohler, Jonas and Bach, Francis and Hofmann, Thomas and Lucchi, Aurelien},
  booktitle = {Advances in Neural Information Processing Systems (NeurIPS)},
  year      = {2020},
}

@inproceedings{feng2022rankdiminishing,
  title     = {Rank Diminishing in Deep Neural Networks},
  author    = {Feng, Ruili and Zheng, Kecheng and Huang, Yukun and Zhao, Deli and Jordan, Michael and Zha, Zheng-Jun},
  booktitle = {Advances in Neural Information Processing Systems (NeurIPS)},
  year      = {2022},
}

@misc{jordan2024muon,
  title        = {Muon: An Optimizer for Hidden Layers in Neural Networks},
  author       = {Jordan, Keller and Jin, Yuchen and Boza, Vlado and You, Jiacheng and Cesista, Franz and Newhouse, Laker and Bernstein, Jeremy},
  year         = {2024},
  howpublished = {\url{https://kellerjordan.github.io/posts/muon/}},
}

@article{liu2025muon,
  title   = {Muon is Scalable for {LLM} Training},
  author  = {Liu, Jingyuan and Su, Jianlin and Yao, Xingcheng and Jiang, Zhejun and Lai, Guokun and Du, Yulun and others},
  journal = {arXiv preprint arXiv:2502.16982},
  year    = {2025},
}

\clearpage
\appendix
\section{Additional Results}
\label{app:additional}

\subsection{Detailed experimental setup}
\label{app:setup-details}

\paragraph{Encoders.}
The five vision encoders in Eagle-X5~\citep{shi2025eagle} are
ConvNeXt-1024~\citep{liu2022convnet},
EVA-02-1024~\citep{fang2023eva02},
CLIP-448~\citep{radford2021learning},
Pix2Struct-1024~\citep{lee2022pix2struct}, and
SAM-1024~\citep{kirillov2023segment}.

\paragraph{Subset enumeration.}
The non-empty subsets comprise 5 singletons, 10 pairs, 10 triples,
5 four-encoder subsets, and the full five-encoder pool. All are
retrained from scratch under an identical recipe.

\paragraph{Benchmark families.}
Following the standard Cambrian-1 partition:
\emph{General} = \{MME-Perception, MMBench-EN, SeedBench, GQA\};
\emph{Knowledge} = \{ScienceQA, MMMU-val, MathVista, AI2D\};
\emph{OCR\&Chart} = \{OCRBench, ChartQA, TextVQA, DocVQA\};
\emph{Vision-Centric} = \{MMVP, RealWorldQA, CV-Bench-2D, CV-Bench-3D\}.
The overall score in the main text is the unweighted average across
all 16 benchmarks.

\subsection{Per-subset training results}
\label{sec:appendix-allpair}

This subsection gives the subset-level results used in the main text.
It includes pair scores and family decompositions
(Table~\ref{tab:allpair}), the full scoreboard for all retrained
subsets (Table~\ref{tab:main_train_ablation}), and the corresponding
parameter counts, throughput, and best-at-$k$ Pareto status
(Table~\ref{tab:compute_pareto}).

\begin{table}[h]
  \centering
  \caption{\textbf{Pair decomposition.} All 10 two encoder subsets,
    sorted by overall score. The highlighted rows compare CLIP+ConvNeXt
    with ConvNeXt+EVA-02. The pair score gap is reported by direction;
    Section~\ref{sec:marginal} focuses on the gap closure pattern
    rather than the absolute gap.}
  \label{tab:allpair}
  \scriptsize
  \resizebox{\linewidth}{!}{\begin{tabular}{l l c c c c c}
\toprule
Pair & Class A $\times$ Class B & Overall & $\Delta$Gen & $\Delta$Know & $\Delta$OCR & $\Delta$Vis \\
\midrule
\rowcolor{black!6}
CLIP+ConvNeXt & Adaptive Complement $\times$ Universal Core & 62.28 & +0.44 & +0.23 & +0.98 & +2.13 \\
\rowcolor{black!6}
ConvNeXt+EVA-02 & Universal Core $\times$ Capacity Specialist & 61.77 & +0.13 & +0.09 & +0.12 & +1.03 \\
ConvNeXt+Pix2Struct & Universal Core $\times$ Niche Specialist & 61.50 & +0.11 & +0.20 & +0.03 & +0.45 \\
ConvNeXt+SAM & Universal Core $\times$ Replaceable & 61.41 & +0.04 & -0.22 & -0.91 & +1.55 \\
CLIP+Pix2Struct & Adaptive Complement $\times$ Niche Specialist & 59.90 & -0.05 & +0.59 & +4.55 & +2.45 \\
Pix2Struct+EVA-02 & Niche Specialist $\times$ Capacity Specialist & 59.66 & +0.47 & +0.64 & +4.06 & +0.75 \\
CLIP+EVA-02 & Adaptive Complement $\times$ Capacity Specialist & 58.87 & +0.00 & +0.39 & +2.29 & +1.31 \\
SAM+EVA-02 & Replaceable $\times$ Capacity Specialist & 57.93 & -0.13 & +0.45 & -0.05 & +2.14 \\
SAM+Pix2Struct & Replaceable $\times$ Niche Specialist & 56.15 & +1.57 & -0.30 & +0.83 & +2.37 \\
CLIP+SAM & Adaptive Complement $\times$ Replaceable & 54.57 & -0.36 & -0.74 & +0.99 & +1.04 \\
\bottomrule
\end{tabular}
}
\end{table}

\FloatBarrier

\begin{table}[p]
  \centering
  \caption{\textbf{Full subset scoreboard.} All non-empty
    encoder subsets trained with the same recipe. Subsets are sorted
    by cardinality and then by overall score, with the best subset at
    each cardinality shown in bold. The table provides a direct lookup
    for each subset's overall and per family scores.}
  \label{tab:main_train_ablation}
  \scriptsize
  \resizebox{\linewidth}{!}{
\begin{tabular}{lccccc c c}
\toprule
Combo & \#enc & General & Knowledge & OCR\&Chart & Vision & Avg & rel. Avg \\
\midrule
\multicolumn{8}{l}{\small\itshape\color{gray}Pool\,size\,5 — 1 combination; rel.\,avg 1.000} \\
\textbf{\texttt{CNSPD}} & 5 & 65.70 & 54.88 & 66.97 & 62.62 & \textbf{62.54} & \textbf{1.000} \\
\midrule
\multicolumn{8}{l}{\small\itshape\color{gray}Pool\,size\,4 — 5 combinations; mean rel.\,avg 0.983, range 0.961–0.995} \\
\textbf{\texttt{CN·PD}} & 4 & 65.48 & 55.19 & 66.44 & 61.76 & \textbf{62.22} & \textbf{0.995} \\
\texttt{·NSPD} & 4 & 65.04 & 55.42 & 66.58 & 60.75 & 61.95 & 0.991 \\
\texttt{CNS·D} & 4 & 65.14 & 54.49 & 65.90 & 60.58 & 61.52 & 0.984 \\
\texttt{CNSP·} & 4 & 64.95 & 54.25 & 66.63 & 60.25 & 61.52 & 0.984 \\
\texttt{C·SPD} & 4 & 64.90 & 55.11 & 61.00 & 59.41 & 60.10 & 0.961 \\
\midrule
\multicolumn{8}{l}{\small\itshape\color{gray}Pool\,size\,3 — 10 combinations; mean rel.\,avg 0.972, range 0.935–0.996} \\
\textbf{\texttt{CN·P·}} & 3 & 65.56 & 55.50 & 67.12 & 60.88 & \textbf{62.26} & \textbf{0.996} \\
\texttt{CN··D} & 3 & 65.58 & 54.79 & 66.39 & 60.91 & 61.92 & 0.990 \\
\texttt{·NSP·} & 3 & 65.27 & 54.82 & 66.65 & 59.93 & 61.67 & 0.986 \\
\texttt{CNS··} & 3 & 64.87 & 54.61 & 66.26 & 60.74 & 61.62 & 0.985 \\
\texttt{·N·PD} & 3 & 65.62 & 54.34 & 65.81 & 60.70 & 61.62 & 0.985 \\
\texttt{·NS·D} & 3 & 64.70 & 54.93 & 65.86 & 60.49 & 61.49 & 0.983 \\
\texttt{C··PD} & 3 & 64.93 & 54.22 & 60.74 & 60.14 & 60.01 & 0.960 \\
\texttt{··SPD} & 3 & 64.01 & 54.01 & 59.41 & 60.84 & 59.57 & 0.953 \\
\texttt{C·SP·} & 3 & 64.08 & 54.26 & 60.17 & 58.86 & 59.34 & 0.949 \\
\texttt{C·S·D} & 3 & 64.83 & 54.65 & 54.10 & 60.34 & 58.48 & 0.935 \\
\midrule
\multicolumn{8}{l}{\small\itshape\color{gray}Pool\,size\,2 — 10 combinations; mean rel.\,avg 0.950, range 0.873–0.996} \\
\textbf{\texttt{CN···}} & 2 & 65.37 & 55.02 & 67.25 & 61.50 & \textbf{62.28} & \textbf{0.996} \\
\texttt{·N··D} & 2 & 65.06 & 54.72 & 66.39 & 60.93 & 61.77 & 0.988 \\
\texttt{·N·P·} & 2 & 65.04 & 54.83 & 66.30 & 59.82 & 61.50 & 0.983 \\
\texttt{·NS··} & 2 & 64.97 & 54.41 & 65.36 & 60.92 & 61.41 & 0.982 \\
\texttt{C··P·} & 2 & 64.76 & 55.38 & 59.94 & 59.52 & 59.90 & 0.958 \\
\texttt{···PD} & 2 & 64.46 & 54.07 & 59.45 & 60.65 & 59.66 & 0.954 \\
\texttt{C···D} & 2 & 64.81 & 55.18 & 54.28 & 61.21 & 58.87 & 0.941 \\
\texttt{··S·D} & 2 & 63.86 & 53.88 & 51.94 & 62.04 & 57.93 & 0.926 \\
\texttt{··SP·} & 2 & 57.86 & 52.54 & 56.22 & 58.00 & 56.15 & 0.898 \\
\texttt{C·S··} & 2 & 64.45 & 54.05 & 41.67 & 58.11 & 54.57 & 0.873 \\
\midrule
\multicolumn{8}{l}{\small\itshape\color{gray}Pool\,size\,1 — 5 combinations; mean rel.\,avg 0.862, range 0.706–0.980} \\
\textbf{\texttt{·N···}} & 1 & 64.93 & 54.63 & 66.27 & 59.37 & \textbf{61.30} & \textbf{0.980} \\
\texttt{····D} & 1 & 63.99 & 53.43 & 51.99 & 59.90 & 57.32 & 0.917 \\
\texttt{C····} & 1 & 64.81 & 54.79 & 40.68 & 57.07 & 54.34 & 0.869 \\
\texttt{···P·} & 1 & 50.10 & 51.90 & 55.39 & 52.87 & 52.56 & 0.840 \\
\texttt{··S··} & 1 & 56.29 & 52.84 & 11.86 & 55.63 & 44.16 & 0.706 \\
\bottomrule
\end{tabular}
}
\end{table}

\begin{table}[p]
  \centering
  \caption{\textbf{Per-subset parameters, throughput, and Pareto
    status.} Throughput in tokens per second, parameter count, and
    relative score for all subsets. The final column marks whether
    a subset lies on the latency score Pareto frontier among the
    best-at-$k$ candidates, using the same selection setting as the
    main text.}
  \label{tab:compute_pareto}
  \scriptsize
  \resizebox{\linewidth}{!}{
\begin{tabular}{lccccccc}
\toprule
Combo & \#Enc & Params (M) & Latency (ms) & Throughput & Avg & rel. Avg & Best-$k$ Pareto \\
\midrule
\texttt{CNSPD} & 5 & 2274 & 427.8 & 2.34 & 62.54 & 1.000 & Yes \\
\texttt{CN·PD} & 4 & 1966 & 362.4 & 2.76 & 62.22 & 0.995 & No \\
\texttt{·NSPD} & 4 & 1970 & 420.4 & 2.38 & 61.95 & 0.991 & No \\
\texttt{CNS·D} & 4 & 1761 & 342.9 & 2.92 & 61.52 & 0.984 & No \\
\texttt{CNSP·} & 4 & 1970 & 367.2 & 2.72 & 61.52 & 0.984 & No \\
\texttt{C·SPD} & 4 & 1427 & 362.2 & 2.76 & 60.10 & 0.961 & No \\
\texttt{CN·P·} & 3 & 1663 & 299.9 & 3.33 & 62.26 & 0.996 & No \\
\texttt{CN··D} & 3 & 1453 & 274.6 & 3.64 & 61.92 & 0.990 & No \\
\texttt{·NSP·} & 3 & 1667 & 359.6 & 2.78 & 61.67 & 0.986 & No \\
\texttt{CNS··} & 3 & 1458 & 279.1 & 3.58 & 61.62 & 0.985 & No \\
\texttt{·N·PD} & 3 & 1663 & 353.7 & 2.83 & 61.62 & 0.985 & No \\
\texttt{·NS·D} & 3 & 1458 & 334.5 & 2.99 & 61.49 & 0.983 & No \\
\texttt{C··PD} & 3 & 1120 & 296.3 & 3.38 & 60.01 & 0.960 & No \\
\texttt{··SPD} & 3 & 1124 & 353.6 & 2.83 & 59.57 & 0.953 & No \\
\texttt{C·SP·} & 3 & 1124 & 334.5 & 2.99 & 59.34 & 0.949 & No \\
\texttt{C·S·D} & 3 & 915 & 274.5 & 3.64 & 58.48 & 0.935 & No \\
\texttt{CN···} & 2 & 1150 & 212.5 & 4.71 & 62.28 & 0.996 & Yes \\
\texttt{·N··D} & 2 & 1150 & 266.8 & 3.75 & 61.77 & 0.988 & No \\
\texttt{·N·P·} & 2 & 1359 & 292.0 & 3.42 & 61.50 & 0.983 & No \\
\texttt{·NS··} & 2 & 1154 & 271.7 & 3.68 & 61.41 & 0.982 & No \\
\texttt{C··P·} & 2 & 816 & 233.0 & 4.29 & 59.90 & 0.958 & No \\
\texttt{···PD} & 2 & 816 & 287.7 & 3.48 & 59.66 & 0.954 & No \\
\texttt{C···D} & 2 & 607 & 207.6 & 4.82 & 58.87 & 0.941 & No \\
\texttt{··S·D} & 2 & 611 & 266.6 & 3.75 & 57.93 & 0.926 & No \\
\texttt{··SP·} & 2 & 820 & 292.6 & 3.42 & 56.15 & 0.898 & No \\
\texttt{C·S··} & 2 & 611 & 212.4 & 4.71 & 54.57 & 0.873 & No \\
\texttt{·N···} & 1 & 846 & 204.6 & 4.89 & 61.30 & 0.980 & Yes \\
\texttt{····D} & 1 & 303 & 199.9 & 5.00 & 57.32 & 0.917 & No \\
\texttt{C····} & 1 & 304 & 145.5 & 6.87 & 54.34 & 0.869 & No \\
\texttt{···P·} & 1 & 513 & 225.5 & 4.43 & 52.56 & 0.840 & No \\
\texttt{··S··} & 1 & 308 & 204.2 & 4.90 & 44.16 & 0.706 & No \\
\bottomrule
\end{tabular}
}
\end{table}

\clearpage

\subsection{Family-conditional Capacity and Necessity profiles}
\label{sec:loo-predictor-diagnostic}

Figure~\ref{fig:main4_capnec}, panels B and C, gives the full
family conditional Cap/Nec decomposition used in
Section~\ref{sec:capnec}. Pooled values from
Table~\ref{tab:taxonomy} are: ConvNeXt (Cap~$0.980$,
Nec~$2.44$\,pp), EVA-02 (Cap~$0.917$, Nec~$1.02$\,pp), CLIP
(Cap~$0.869$, Nec~$0.59$\,pp), Pix2Struct (Cap~$0.840$,
Nec~$1.02$\,pp), and SAM (Cap~$0.706$, Nec~$0.32$\,pp).

The family conditional view separates encoders that look similar in
the pooled numbers. Pix2Struct has much higher OCR\&Chart Capacity
than General Capacity, matching its role as a niche specialist.
EVA-02 has the highest Vision-Centric Capacity, making it a strong alternative anchor for
vision heavy settings. Knowledge Capacity is high for all encoders
(SAM reaches $0.967$, panel B), reflecting the strong LLM prior on
that family. For this reason, Section~\ref{sec:capnec} does not use
Knowledge when drawing family conditional conclusions.

For Necessity, panel C shows that removing ConvNeXt causes the largest
single encoder drop on OCR\&Chart ($5.97$\,pp). SAM is close to zero
across all families, with pooled Necessity at most $0.32$\,pp, so it
is the natural first encoder to remove when the pool must be reduced.

\subsection{Effective rank by subset size}
\label{sec:appendix-rank-by-pool}

Table~\ref{tab:rank-vs-pool} reports the mean effective rank of each
encoder at each subset size $k$. Section~\ref{sec:mechanism} uses
these numbers in two places. First, the $k{=}2$ column supports the
claim in Section~\ref{sec:rank_synthesis} that CLIP's mean rank rises
by $+9.9$ dims across its four heterogeneous pairings
(Table~\ref{tab:rank-vs-pool}: $64.7 - 54.8$). CLIP is the only
encoder whose rank increases when it is paired with another encoder.
Second, the variation across encoders within the same $k$ gives the
within $k$ centred residual diagnostic used in the footnote of
Section~\ref{sec:rank_vs_params}. From $k{=}1$ to $k{=}5$, only CLIP
has a positive change ($+12$ dims). The other four encoders lose
effective rank as more encoders share the projector's fixed output
budget.

\begin{table}[!htbp]
  \centering
  \caption{\textbf{Mean per encoder effective rank by subset size
    $k$.} Each cell gives the mean effective rank of the named encoder,
    averaged over all subsets that contain that encoder at the given
    $k$. Standard deviation is shown where applicable. The last column
    reports the change from $k{=}1$ to $k{=}5$.}
  \label{tab:rank-vs-pool}
  \small
  \begin{tabular}{lcccccc}
\toprule
Encoder & $k{=}1$ & $k{=}2$ & $k{=}3$ & $k{=}4$ & $k{=}5$ & $\Delta_{1\to5}$\\
\midrule
CLIP & $54.8$ & $64.7_{\pm3.7}$ & $62.6_{\pm3.3}$ & $64.7_{\pm5.5}$ & $66.9$ & $\uparrow$\,$+12$\\
ConvNeXt & $54.0$ & $54.9_{\pm11.9}$ & $52.0_{\pm20.9}$ & $63.6_{\pm12.4}$ & $27.5$ & $\downarrow$\,$-26$\\
SAM & $17.5$ & $11.7_{\pm3.7}$ & $8.3_{\pm2.2}$ & $7.5_{\pm1.5}$ & $5.6$ & $\downarrow$\,$-12$\\
Pix2Struct & $39.3$ & $29.8_{\pm7.2}$ & $29.5_{\pm3.0}$ & $27.7_{\pm1.7}$ & $26.6$ & $\downarrow$\,$-13$\\
EVA-02 & $44.0$ & $40.9_{\pm4.2}$ & $40.1_{\pm5.5}$ & $36.0_{\pm6.6}$ & $33.4$ & $\downarrow$\,$-11$\\
\midrule
\multicolumn{1}{l}{\emph{models per column}} & $n{=}5$ & $n{=}10$ & $n{=}10$ & $n{=}5$ & $n{=}1$ & --\\
\bottomrule
\end{tabular}

\end{table}

\FloatBarrier

\subsection{Sample-level consistency checks}
\label{sec:appendix-extra-diagnostics}

\paragraph{Drift-to-correctness (CKA).}
For each encoder in the five encoder full pool, we compute a
per-sample CKA score between the encoder's pre-projector activations
after fine tuning and its activations at the pre-training checkpoint.
We then compare this drift measure between correctly and incorrectly
answered samples using Welch's $t$-test and Cohen's~$d$
(Section~\ref{sec:rank_synthesis}).

For the full pool baseline ($n_{\text{correct}}{=}17{,}470$;
$n_{\text{incorrect}}{=}8{,}202$), CLIP and EVA-02 have negative $d$
values (CLIP: $d{=}{-}0.165$; EVA-02: $d{=}{-}0.131$; both
$p$ below $0.001$). Correctly answered queries are associated with
lower deviation from the pretrained representation. This matches the
view that these encoders provide stable, pretrain aligned visual
grounding on the samples the model answers correctly. Pix2Struct and
SAM have positive $d$ values (Pix2Struct: $d{=}{+}0.170$; SAM:
$d{=}{+}0.121$; both $p$ below $0.001$). For these encoders, correct
answers are associated with larger drift, suggesting that fine tuning
moves their representations toward document and boundary cues that are
useful for the answer. ConvNeXt is much smaller
($d{=}{+}0.026$; $p{=}0.042$), in line with its role as a
complementary anchor rather than the main driver of this sample level
CKA signal.

\paragraph{IM-flip negative control.}
For each encoder, family, and variant cell, we compute the Pearson
correlation between two quantities: the representation divergence
caused by ablation, measured as the L2 drift norm at the projector
input when the encoder's input channel is zeroed under IM, and the
sample level flip rate under IM. This measures how much the encoder's
own channel slice shifts when its input is masked. It is separate from
the drift to correctness measure above, which compares pre-training
and fine-tuning checkpoints using CKA.

Across the 80 cells (5 encoders $\times$ 4 families $\times$
4 variants), the maximum absolute Pearson correlation is $0.105$, and
the mean absolute correlation is $0.030$. Thus, ablation induced
representation divergence and IM flip rate are nearly uncorrelated at
the sample level. The two measurements are capturing different parts
of encoder behaviour.

\subsection{CKA convergence with encoder-pool size}
\label{sec:appendix-cka-convergence}

Table~\ref{tab:cka-convergence} shows how projector layer CKA between
two retrained models changes with the number of encoders they share.
Across all 16{,}896 ordered model pairs, the Spearman correlation
between projector CKA and the minimum shared encoder count is
$\rho{=}0.202$ ($p$ below $0.001$). Mean projector CKA increases with
pool overlap: $0.754$ for pairs sharing one encoder, $0.800$ for two,
$0.830$ for three, and $0.852$ for four. The same pattern appears
within each task family
($\rho_{\text{General}}{=}0.299$;
$\rho_{\text{OCR\&Chart}}{=}0.231$;
$\rho_{\text{Vision-Centric}}{=}0.242$;
$\rho_{\text{Knowledge}}{=}0.086$). This suggests that the projector
geometry is tied to which encoders are present, rather than being
explained only by random initialisation.

\begin{table}[!htbp]
  \centering
  \caption{\textbf{Projector CKA by shared encoder count.}
    Spearman $\rho$ with 95\% confidence interval between projector
    CKA and minimum shared encoder count, together with mean projector
    CKA for each shared count bucket, grouped by task family.}
  \label{tab:cka-convergence}
  \small
  \resizebox{\linewidth}{!}{
\begin{tabular}{lrrrrrrr}
\toprule
Family & Shared-1 & Shared-2 & Shared-3 & Shared-4 & Mixed & $\rho$ \\
\midrule
Overall         & 0.754 & 0.800 & 0.830 & 0.852 & 0.814 & 0.202 \\
General         & 0.759 & 0.793 & 0.838 & 0.862 & 0.809 & 0.299 \\
Knowledge       & 0.746 & 0.772 & 0.774 & 0.808 & 0.788 & 0.086 \\
OCR \& Chart    & 0.699 & 0.775 & 0.806 & 0.829 & 0.789 & 0.231 \\
Vision-Centric  & 0.810 & 0.861 & 0.901 & 0.908 & 0.870 & 0.242 \\
\bottomrule
\end{tabular}
}
\end{table}

\FloatBarrier

\subsection{Partial Information Decomposition between encoder pairs}
\label{sec:appendix-pid}

Table~\ref{tab:pid} reports a proxy Partial Information Decomposition
(PID) between encoder pairs. We decompose the change in projector CKA
when an encoder is added to a singleton pool into three terms: shared
information, unique information from each encoder, and synergy. Here,
shared information is the CKA already captured by the existing
singleton, while synergy measures the interaction gain from using the
pair.

Across all task family cells, the shared term is the largest component
(about $0.55$ to $0.85$ of total CKA). Unique contributions are small,
around $0.02$ or lower, and synergy is close to zero. At the CKA
level, pairwise complementarity in Eagle-X5 therefore looks mostly
additive rather than strongly synergistic. This agrees with the rank
budget view: encoders contribute largely independent subspace slices,
and the projector combines them without needing a large cross encoder
interaction term in the representation geometry.

\begin{table}[!htbp]
  \centering
  \caption{\textbf{Proxy PID shared CKA between encoder pairs.}
    Mean \texttt{proxy\_shared\_cka}, defined as the CKA between the
    two encoders' pre-projector representations, aggregated by
    benchmark family. All values pair CLIP with one partner encoder.
    Unique and synergy terms are close to zero for percentage scaled
    tasks, around $0.02$ or lower, and are omitted for readability.}
  \label{tab:pid}
  \scriptsize
  \resizebox{\linewidth}{!}{
\begin{tabular}{lrrrr}
\toprule
Pair (CLIP $+$ X) & General & Knowledge & OCR \& Chart & Vision-Centric \\
\midrule
CLIP $+$ ConvNeXt   & 0.848 & 0.869 & 0.783 & 0.873 \\
CLIP $+$ EVA-02     & 0.844 & 0.848 & 0.824 & 0.877 \\
CLIP $+$ Pix2Struct & 0.661 & 0.648 & 0.647 & 0.738 \\
CLIP $+$ SAM        & 0.768 & 0.739 & 0.655 & 0.815 \\
\bottomrule
\end{tabular}
}
\end{table}

\FloatBarrier

\subsection{Scope and limitations}
\label{sec:appendix-scope-limitations}

Findings are calibrated to channel-concatenation fusion, the
Vicuna-7B decoder, and a single five-encoder pool; the rank-budget
argument relies on a shared projector with fixed output dimension
and does not transfer directly to cross-attention or routing
architectures, where each encoder has its own pathway. The
five-class taxonomy is a descriptive label for Eagle-X5; whether
the Universal Core and Adaptive Complement pattern generalises
beyond CLIP is not settled by our data. Each subset is trained
once, so small per-family differences are directional evidence
rather than significance claims. Effects of decoder size on the
IM/TR discrepancy and on the saturation point, as well as
extensions to larger encoder pools, are left to future work.

\end{document}